\documentclass[10pt, conference]{IEEEtran}
\IEEEoverridecommandlockouts
\usepackage{amsmath,amssymb,amsfonts,amsthm}
\usepackage{graphicx}
\usepackage{svg}
\usepackage{caption}
\usepackage{algorithm}
\usepackage{algpseudocode}
\usepackage{textcomp}
\usepackage{xcolor}
\usepackage{tikz}
\usetikzlibrary{shapes.geometric, arrows.meta, positioning, fit, calc, shadows}
\usepackage{array}
\usepackage{booktabs}
\usepackage[backend=biber,style=ieee,sorting=none]{biblatex}
\AtEveryBibitem{
  \clearfield{url}\clearfield{doi}
  \clearfield{urldate}\clearfield{eprint}
}

\makeatletter
\newcommand{\linebreakand}{%
  \end{@IEEEauthorhalign}\\
  \begin{@IEEEauthorhalign}
}
\makeatother

\begin{document}

\title{QScheduler: Adaptive Gradient Sampling for Zeroth-Order On-Device Training on INT8 NPUs}

\author{
\IEEEauthorblockN{%
  Victor Felipe Domingues do Amaral$^{\dagger\ddagger}$,
  Pierre Demaj$^{\dagger}$,
  Erwan Libessart$^{\ddagger}$\\
  Laurent Folliot$^{\dagger}$,
  Anthony Kolar$^{\ddagger}$,
  Philippe Bénabès$^{\ddagger}$%
}
\IEEEauthorblockA{%
  $^{\dagger}$ STMicroelectronics, France\\
  \{victor.dominguesdoamaral, pierre.demaj, laurent.folliot\}@st.com\\
  $^{\ddagger}$ GeePs, CNRS, CentraleSupélec, Université Paris-Saclay, Sorbonne Université, France\\
  \{victor.domingues, erwan.libessart, anthony.kolar, philippe.benabes\}@centralesupelec.fr
}
}

\maketitle

\begin{tikzpicture}[remember picture,overlay]
  \node[anchor=south,yshift=18pt] at (current page.south)
  {\parbox{\textwidth}{\footnotesize \copyright~2026 IEEE. Personal use of this material is permitted. Permission from IEEE must be obtained for all other uses, in any current or future media, including reprinting/republishing this material for advertising or promotional purposes, creating new collective works, for resale or redistribution to servers or lists, or reuse of any copyrighted component of this work in other works.}};
\end{tikzpicture}

\begin{abstract}
Zeroth-Order (ZO) optimization enables On-Device Learning (ODL) on NPU-equipped microcontrollers by estimating gradients through forward passes alone, bypassing the need for backpropagation primitives and reducing memory requirements. The number of gradient samples $q$ critically affects training: insufficient samples produce noisy gradients that plateau early, while excessive samples consume more computational resources. However, finding an optimal $q$ typically requires costly hyperparameter searches. This work introduces QScheduler, an adaptive algorithm that adjusts $q$ based on training progress, and provides the first proof-of-concept of INT8 quantized on-device training on the STM32N6's Neural-ART NPU. Experiments on EuroSAT and STL-10 show that QScheduler matches well-tuned fixed-$q$ configurations for both ResNet18 and MobileNetV2, without requiring prior $q$ hyperparameter optimization.
\end{abstract}

\begin{IEEEkeywords}
On-Device Learning, Zeroth-Order Optimization, INT8 Training, Transfer Learning, Neural Processing Unit
\end{IEEEkeywords}

\section{Introduction}

Edge AI brings machine learning inference to resource-constrained devices such as microcontrollers (MCUs) and systems-on-chip (SoCs) equipped with neural processing units (NPUs). Unlike cloud or mobile platforms, these embedded devices operate under strict memory, compute, and energy budgets. A typical microcontroller may have only a few hundred kilobytes of SRAM and a few megabytes of Flash storage, orders of magnitude less than mobile or cloud systems \cite{lin_mcunet_2020}. Despite these constraints, recent design efforts across models, compilers, and hardware have enabled practical on-device inference for vision, audio, and sensor applications \cite{gholami_survey_2021, lin_mcunet_2020, stm32modelzoo, cai_tinytl_2021,lin_mcunetv2_2021}.

On-Device Learning (ODL) extends this paradigm beyond inference to enable training or fine-tuning models directly on embedded devices using locally collected data. ODL is increasingly important for three key reasons: (i) \textit{Privacy}: sensitive data remains on-device, avoiding cloud transmission and simplifying compliance; (ii) \textit{Connectivity and latency}: many deployments face intermittent networks, and local adaptation eliminates round-trip delays; and (iii) \textit{Energy efficiency}: wireless transmission can dominate power budgets, and local incremental updates can consume less energy than repeated data transfers \cite{ren_tinyol_2021}. ODL thus opens the door to personalized, adaptive AI models that evolve continuously on the edge.

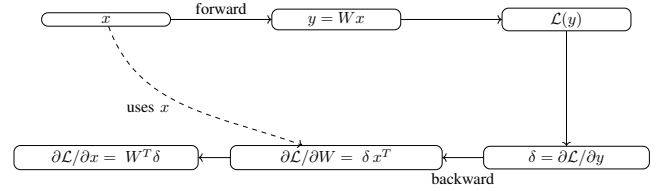
\begin{figure}[t]
\centering
        \resizebox{0.95\columnwidth}{!}{%
        \begin{tikzpicture}
            \node[draw, rounded corners, align=center, inner sep=2pt, minimum width=2.8cm] (x) at (0,0) {$x$};
            \node[draw, rounded corners, align=center, inner sep=2pt, minimum width=2.8cm] (wx) at (5,0) {$y = W x$};
            \node[draw, rounded corners, align=center, inner sep=2pt, minimum width=2.8cm] (loss) at (10,0) {$\mathcal{L}(y)$};
            \draw[->] (x) -- node[above]{forward} (wx);
            \draw[->] (wx) -- (loss);

            \node[draw, rounded corners, align=center, inner sep=2pt, minimum width=3.6cm] (delta) at (10,-3) {$\delta = \partial \mathcal{L}/\partial y$};
            \node[draw, rounded corners, align=center, inner sep=2pt, minimum width=4.6cm] (dW) at (5,-3) {$\partial \mathcal{L}/\partial W =\; \delta\, x^{T}$};
            \node[draw, rounded corners, align=center, inner sep=2pt, minimum width=4.0cm] (dx) at (0,-3) {$\partial \mathcal{L}/\partial x =\; W^{T} \delta$};
            \draw[->] (loss) -- (delta);
            \draw[->] (delta) -- node[pos=0.5, below=6pt]{backward} (dW);
            \draw[->] (dW) -- (dx);

            \draw[->, dashed] (x) to [out=-70, in=160] node[pos=0.3, below=10pt]{uses $x$} (dW);
        \end{tikzpicture}%
        }
\caption{Backpropagation memory dependency: the backward pass requires stored activations from the forward pass (dashed arrow).}
\label{fig:bp_dependency}
\end{figure}

However, the dominant training approach, backpropagation (BP), poses challenges for embedded SoCs. BP requires a backward pass that propagates error gradients through the network to update parameters \cite{rumelhart_backprop_1986}. As shown in Figure~\ref{fig:bp_dependency}, computing weight gradients depends on both the error signal $\delta$ and stored activations $x$ from the forward pass.

This backward pass requires computational primitives such as gradient accumulation and transposed convolutions that many embedded NPUs do not fully support, as they are optimized for inference throughput \cite{feng_stepping_2024}. Memory-reduction techniques like activation checkpointing \cite{chen_training_2016, patil_poet_2022} address memory constraints but still require backward kernels unavailable on hardware such as integer-only NPUs.

An alternative strategy is to eliminate the backward pass entirely using Zeroth-Order (ZO) optimization, which estimates gradients using only forward evaluations of the loss function \cite{nesterov_random_2017, spall_spsa_1992}. This forward-only formulation naturally aligns with inference-optimized accelerators while also reducing memory requirements, which is crucial for on-device learning. However, ZO gradient estimates are inherently noisy, typically underperforming backpropagation in convergence speed and final accuracy.

In this paper, we propose an adaptive framework for ZO-based on-device learning. Our key contributions are:
\begin{itemize}
    \item \textit{QScheduler}, a runtime algorithm that monitors training progress and adaptively increases the number of gradient samples $q$ only when the model plateaus, eliminating the need for costly hyperparameter searches to determine the optimal $q$.
    \item An implementation of this framework for INT8 quantized on-device training, validated on the STM32N6 microcontroller and its Neural-ART accelerator \cite{stm32n6_datasheet}.
\end{itemize}

\section{Related Work}

\subsection{Backpropagation on Embedded Devices}

Several approaches address ODL within these constraints. TinyTL \cite{cai_tinytl_2021} reduces memory by training only bias terms and a small adapter module, avoiding full activation storage. TinyOL \cite{ren_tinyol_2021} proposes lightweight architectures designed for on-device updates. Sparse update methods \cite{lin_-device_2022} reduce memory by training only a subset of parameters. However, these methods still assume backward-pass support on the target hardware. For NPUs adapted only for forward inference (common in integer-only accelerators), alternative training strategies are necessary.

\subsection{Zeroth-Order Optimization for Neural Networks}

Zeroth-Order (ZO) optimization estimates gradients using only forward evaluations of the loss function, without explicit gradient computation \cite{nesterov_random_2017, spall_spsa_1992}. Given a parameter vector $\theta$ and a loss $\mathcal{L}(\theta)$, a basic ZO random estimator perturbs $\theta$ by a small random direction $z$ and approximates the gradient via finite differences:
\begin{equation}
\widehat{\nabla}\,\mathcal{L}(\theta; z) \;=\; \frac{\mathcal{L}(\theta + \mu\,z) - \mathcal{L}(\theta)}{\mu}\; z,
\label{eq:zo_estimator}
\end{equation}
where $\mu > 0$ is the perturbation radius and $z$ is typically sampled from $\mathcal{N}(0,I)$. This estimator requires only two forward passes per update and no backward kernels.

A key limitation of this estimator is its high variance, which scales with parameter dimension \cite{nesterov_random_2017}. A standard variance-reduction technique averages over $q$ independent perturbations:
\begin{equation}
\widehat{\nabla}\,\mathcal{L}_q(\theta) \;=\; \frac{1}{q}\sum_{i=1}^{q} \widehat{\nabla}\,\mathcal{L}(\theta; z_i),
\label{eq:multi_sample}
\end{equation}
reducing variance by $\mathcal{O}(1/q)$ at the cost of $q + 1$ forward passes per update. The perturbation radius $\mu$ controls a bias-variance trade-off: larger $\mu$ introduces bias, while smaller $\mu$ increases numerical instability (typical values: $10^{-5}$ to $10^{-3}$). However, selecting the optimal $q$ remains an open problem.

MeZO \cite{malladi2023mezo} demonstrates that ZO optimization can fine-tune large language models with memory consumption comparable to inference, using a seed-based approach to regenerate perturbations on-the-fly. For embedded systems, Stepping Forward on the Last Mile \cite{feng_stepping_2024} demonstrates ZO-based fine-tuning with quantized models on edge devices. These works typically use a fixed number of gradient samples $q$; our work addresses the challenge of selecting $q$ adaptively during training.

\section{Methodology}

We first introduce QScheduler, our adaptive algorithm for gradient sampling, and then detail the INT8 training framework targeting the STM32N6 microcontroller.

\subsection{Adaptive Gradient Sampling: QScheduler}

We propose QScheduler (Algorithm~\ref{alg:qscheduler}), which monitors training progress and adaptively increases $q$ only when improvement stalls. The algorithm tracks a smoothed best metric $M_{\text{best}}$ (e.g., validation loss or accuracy) and counts consecutive epochs $C_{\text{epochs}}$ without significant improvement (defined by tolerance $\tau$). We define the preference relation $M \succ M_{\text{best}}$ to indicate that metric $M$ is preferable to $M_{\text{best}}$ by at least $\tau$ (e.g., $M < M_{\text{best}} - \tau$ for loss, or $M > M_{\text{best}} + \tau$ for accuracy). When the patience threshold is exceeded, $q$ is multiplied by $\alpha$ up to a maximum $Q_{\text{max}}$, and the counter resets. We restrict $q$ to a discrete set (e.g., $\{2, 4, 8, ..., Q_{\text{max}}\}$).

\begin{algorithm}[h]
\caption{QScheduler}
\label{alg:qscheduler}
\begin{algorithmic}[1]
\Require patience $p \in \mathbb{N}^+$, multiplier $\alpha > 1$, initial samples $q_0$, maximum samples $q_{\text{max}}$
\State \textbf{Initialize:} $q \leftarrow q_0$,\quad $M_{\text{best}} \leftarrow \varnothing$,\quad counter $\leftarrow 0$
\vspace{0.3em}
\Function{Step}{metric $M$}
  \If{$M_{\text{best}} = \varnothing$ \textbf{or} $M \succ M_{\text{best}}$}
    \State $M_{\text{best}} \leftarrow M$;\quad counter $\leftarrow 0$
  \Else
    \State counter $\leftarrow$ counter $+ 1$
  \EndIf
  \If{counter $> p$}
    \State $q \leftarrow \min(\alpha \cdot q,\ q_{\text{max}})$;\quad counter $\leftarrow 0$
  \EndIf
  \State \textbf{return} $q$
\EndFunction
\end{algorithmic}
\end{algorithm}

The optimal $q$ is problem-dependent and unknown a priori---too few samples yield noisy gradients that plateau early, while too many waste computation. QScheduler addresses this uncertainty by starting with a small $q$ and increasing it only when progress stalls, allocating computational budget to gradient quality on demand. The patience-based monitoring is inspired by ReduceLROnPlateau \cite{pytorch_reduce_on_plateau}, but acts on a different dimension: instead of reducing step size, it increases $q$ to reduce estimation variance. The per-step cost scales as:

\begin{equation}
C_{\text{ZO}}(q) \;\approx\; q\,(C_{\text{fwd}} + C_{\text{perturb}}),
\label{eq:compute_cost}
\end{equation}
where $C_{\text{fwd}}$ is the computational cost of one forward pass and $C_{\text{perturb}}$ is the cost of perturbing the set of parameters, since each estimation requires perturbing the parameters.

\subsection{ZO Training on Quantized NPUs}

We target SoC hardware featuring quantized NPUs operating on INT8 arithmetic. While we validate on the STM32N6 microcontroller, the approach applies to similar embedded systems with integer-only inference accelerators lacking backward-pass support.

\begin{figure}[t]
\centering
\includegraphics[trim=1mm 1mm 1mm 1mm, clip, width=0.95\columnwidth, height=5.5cm, keepaspectratio=false]{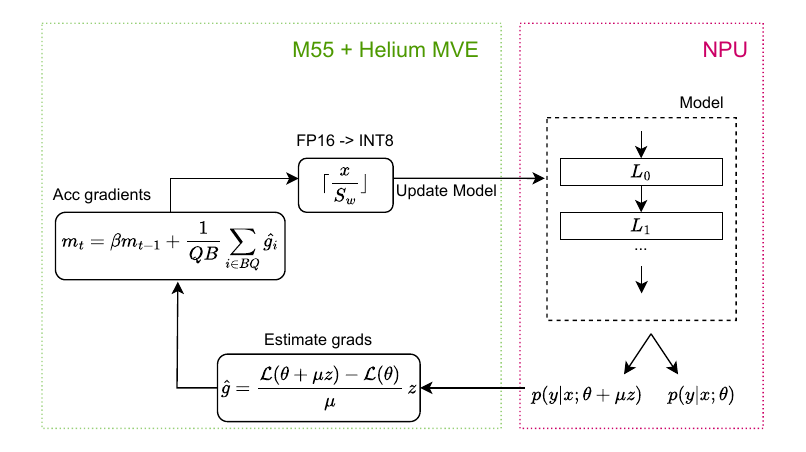}
\caption{ZO training framework showing the computational partition between CPU and NPU. The CPU performs gradient estimation, accumulation, and weight updates. The NPU executes only forward passes.}
\label{fig:zo_framework}
\end{figure}

Figure~\ref{fig:zo_framework} illustrates the division of labor between CPU and NPU during ZO training. The NPU executes quantized forward passes to compute output distributions for both original and perturbed models. The CPU computes losses $\mathcal{L}(\theta)$ and $\mathcal{L}(\theta + \mu\,z)$, generates perturbation vectors $z$, estimates gradients via Eq.~\eqref{eq:zo_estimator}, accumulates $q$ samples per Eq.~\eqref{eq:multi_sample}, and applies weight updates.

Perturbations are sampled from a Rademacher distribution ($z_i \in \{-1, +1\}$) and applied directly in INT8: $w_{\text{perturbed}} = w_{\text{INT8}} + z$. This integer-space perturbation $w_{\text{INT8}} \pm 1$ corresponds to $\theta \pm S_w$ in real space, where $S_w$ is the quantization scale. A constraint arises from the discrete nature of quantized weights: if $\mu < S_w$, the perturbation is insufficient to change the quantized value, yielding zero gradient estimates. This necessitates $\mu \geq S_w$ \cite{feng_stepping_2024}.

To stabilize training under noisy ZO estimates, we use a gradient accumulation buffer with momentum. Momentum is widely used in optimization \cite{kingma_adam_2015} and maintains an exponentially weighted accumulation of past gradients ($m_t \gets \beta m_{t-1} + \widehat{g}_t$), dampening oscillations from high-variance samples.

\subsection{Experimental Setup}

The framework is evaluated on two datasets: EuroSAT \cite{helber_eurosat_2019} for simulation experiments and STL-10 \cite{coates_stl10_2011} for on-device validation.

We use ResNet18 and MobileNetV2 backbones pre-trained on ImageNet, fine-tuning the last fully-connected layer (transfer learning). Training follows standard practices: data augmentation includes random horizontal and vertical flips with random cropping, and the train/validation split is 80\%/20\%. The learning rate follows a cosine decay schedule. Early stopping terminates training if validation accuracy stagnates for 30 epochs or if overfitting is detected. Hyperparameters were optimized via Bayesian search \cite{snoek_bayesian_2012}; Table~\ref{tab:hyperparam_search} lists the search space.

\begin{table}[h]
\centering
\caption{Hyperparameter search space for Bayesian optimization.}
\label{tab:hyperparam_search}
\begin{tabular}{ll}
\toprule
\textbf{Parameter} & \textbf{Search Space} \\
\midrule
Learning rate & $\{10^{-4}, 5{\times}10^{-4}, 10^{-3}, 5{\times}10^{-3}\}$ \\
Momentum $\beta$ & $\{0.90, 0.95, 0.99\}$ \\
Perturbation $\mu$ & $\{10^{-5}, 5{\times}10^{-5}, 10^{-4}, 5{\times}10^{-4}, 10^{-3}\}$ \\
Batch size & $\{8, 16, 32, 64\}$ \\
\bottomrule
\end{tabular}
\end{table}

For INT8 simulation experiments, we use PyTorch with fake quantization (INT8 with per-channel scales) to emulate integer-space updates on a host machine. Training follows a three-stage procedure: (1) a warmup phase using floating-point training until the model reaches around 30\% accuracy; (2) calibration of quantization scales using 25 batches of training data; and (3) quantized ZO training with integer-space perturbations. Warmup before quantized training is a common practice to avoid poorly calibrated initializations \cite{lin_-device_2022}.

\subsection{On-Device Implementation}

For hardware validation, we deploy on the STM32N6 microcontroller featuring an ARM Cortex-M55 CPU and Neural-ART NPU. Unlike the PyTorch simulations, the NPU executes real INT8 arithmetic. Since the NPU lacks backward-pass support, it is an ideal target for our ZO approach. Model weights and gradient buffers reside in external PSRAM, while internal SRAM holds activations during inference. At initialization, trainable weights are copied from flash to PSRAM, as only PSRAM is memory-mapped for read/write access.

As in simulation, we first perform warmup and calibration in PyTorch. For deployment, we export the calibrated model to ONNX and compile it using STEdgeAI \cite{st_edge_ai}, then extend the generated inference code with our ZO training framework implemented in C and compiled with Arm Compiler \cite{arm_compiler}. At runtime, the on-device procedure is:
\begin{itemize}
    \item \textbf{Boot:} Initialize hardware peripherals (PSRAM, external flash, NPU). Copy trainable layer weights from flash to PSRAM, which provides the read/write access required for weight updates. Allocate gradient accumulation buffer and momentum state in PSRAM. Configure the dataloader to stream training samples from external flash.
    \item \textbf{Training loop:} For each batch, run a forward pass on the NPU to compute the baseline loss $\mathcal{L}(\theta)$. Then, for each of $q$ perturbation samples, apply a Rademacher perturbation $z_i \in \{-1,+1\}^d$ to the weights, run another forward pass to obtain $\mathcal{L}(\theta + \mu z_i)$, and accumulate the gradient estimate. After processing all $q$ samples, update weights using momentum SGD. At each validation epoch, QScheduler monitors accuracy: if progress stagnates for a defined patience period, $q$ is multiplied by factor $\alpha$ up to $Q_{\max}$.
\end{itemize}

\section{Results}

We evaluate our ZO framework and QScheduler in two settings: (i) simulation experiments in PyTorch to compare accuracy across models and quantization levels, and (ii) on-device experiment on the STM32N6 to validate real hardware deployment.

\subsection{Simulation Results}

Using the experimental setup described in Section~III-C, we compare fixed-$q$ baselines ($q \in \{4, 8, 16, 32, ...\}$) against QScheduler, testing both floating-point and INT8 quantized models.

Table~\ref{tab:qscheduler_params} lists the QScheduler parameters. We start with a small $q_0$ to benefit from fast initial progress, then allow the scheduler to increase $q$ by factor $\alpha$ when validation accuracy stagnates for \textit{patience} epochs. We tested $\alpha \in \{1.5, 2.0\}$ and patience $\in \{5, 10\}$ epochs, finding that $\alpha=2$ and patience$=5$ yielded the best results in our experiments. $Q_{\text{max}}$ caps the maximum samples to avoid excessive computation in late training.

\begin{table}[h]
\centering
\caption{QScheduler hyperparameters.}
\label{tab:qscheduler_params}
\begin{tabular}{lcc}
\toprule
\textbf{Parameter} & \textbf{Symbol} & \textbf{Value} \\
\midrule
Initial samples & $q_0$ & 8 \\
Multiplier & $\alpha$ & 2.0 \\
Patience (epochs) & p & 5 \\
Maximum samples & $Q_{\text{max}}$ & \{64,1024\} \\
\bottomrule
\end{tabular}
\end{table}

For INT8 models, we follow the three-stage training procedure described in Section~III-C (warmup, calibration, and quantized training).

Tables~\ref{tab:accuracy_resnet18} and~\ref{tab:accuracy_mobilenetv2} summarize final validation accuracy, computed as the mean over the last 10 epochs of each run (mean $\pm$ standard deviation over 5 runs). QScheduler achieves accuracy comparable to well-tuned fixed-$q$ configurations without requiring prior knowledge of the optimal $q$. INT8 quantized training incurs a significant accuracy drop compared to floating-point.

\begin{table}[h]
\centering
\begin{tabular}{lcc}
\toprule
\textbf{Method} & \textbf{Float} & \textbf{INT8} \\
\midrule
$q=4$ & 75.46 $\pm$ 3.45 & 28.72 $\pm$ 2.05 \\
$q=8$ & 81.73 $\pm$ 1.91 & 36.09 $\pm$ 2.40 \\
$q=16$ & 85.75 $\pm$ 1.17 & 49.87 $\pm$ 1.17 \\
$q=32$ & 87.80 $\pm$ 0.72 & 62.39 $\pm$ 0.97 \\
$q=64$ & 89.63 $\pm$ 0.64 & 72.97 $\pm$ 1.10 \\
$q=256$ & 90.25 $\pm$ 0.50 & 83.23 $\pm$ 0.23 \\
$q=512$ & 90.44 $\pm$ 0.45 & 84.86 $\pm$ 0.12 \\
$q=1024$ & 90.87 $\pm$ 0.22 & 86.05 $\pm$ 0.18 \\
\midrule
QScheduler & 90.48 $\pm$ 0.36 & 86.21 $\pm$ 0.20 \\
\bottomrule
\end{tabular}
\caption{ResNet18 validation accuracy (\%) on EuroSAT ($Q_{max}$ = 1024).}
\label{tab:accuracy_resnet18}
\end{table}

\begin{table}[h]
\centering
\begin{tabular}{lcc}
\toprule
\textbf{Method} & \textbf{Float} & \textbf{INT8} \\
\midrule
$q=4$ & 65.49 $\pm$ 4.12 & 54.76 $\pm$ 2.33 \\
$q=8$ & 81.04 $\pm$ 1.36 & 69.80 $\pm$ 1.19 \\
$q=16$ & 86.45 $\pm$ 0.56 & 79.15 $\pm$ 1.23 \\
$q=32$ & 89.38 $\pm$ 0.77 & 84.40 $\pm$ 0.94 \\
$q=64$ & 91.73 $\pm$ 0.55 & 85.40 $\pm$ 0.67 \\
\midrule
QScheduler & 91.61 $\pm$ 0.18 & 85.09 $\pm$ 0.38 \\
\bottomrule
\end{tabular}
\caption{MobileNetV2 validation accuracy (\%) on EuroSAT ($Q_{max}$ = 64).}
\label{tab:accuracy_mobilenetv2}
\end{table}

\begin{figure}[t]
\centering
\includegraphics[width=\columnwidth]{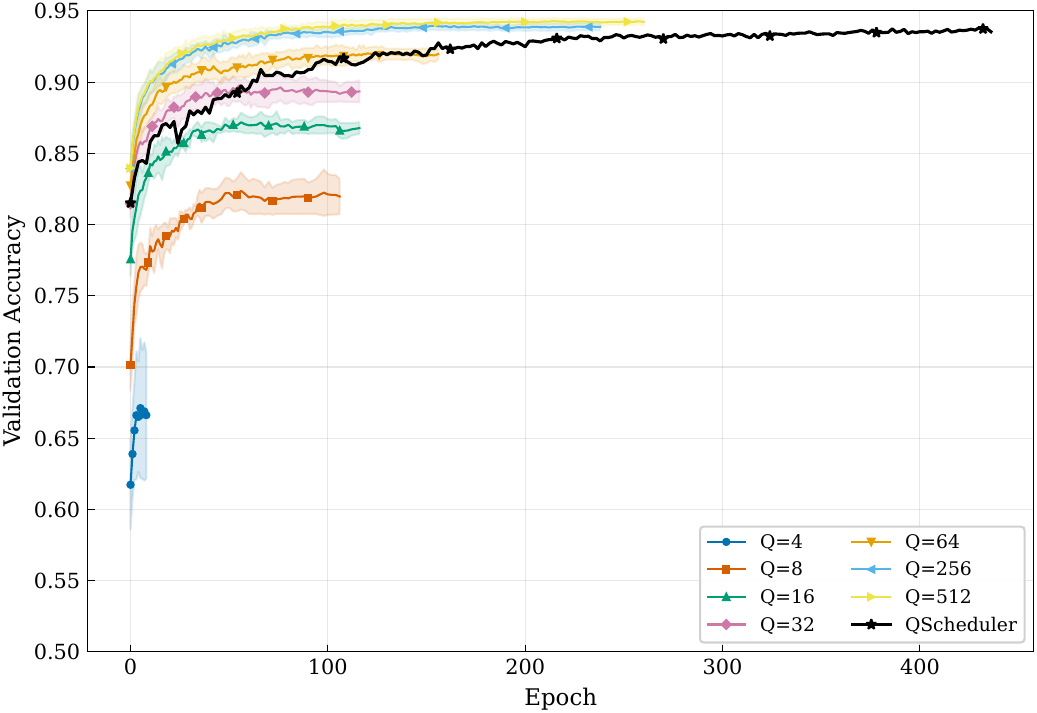}\\[2mm]
\includegraphics[width=\columnwidth]{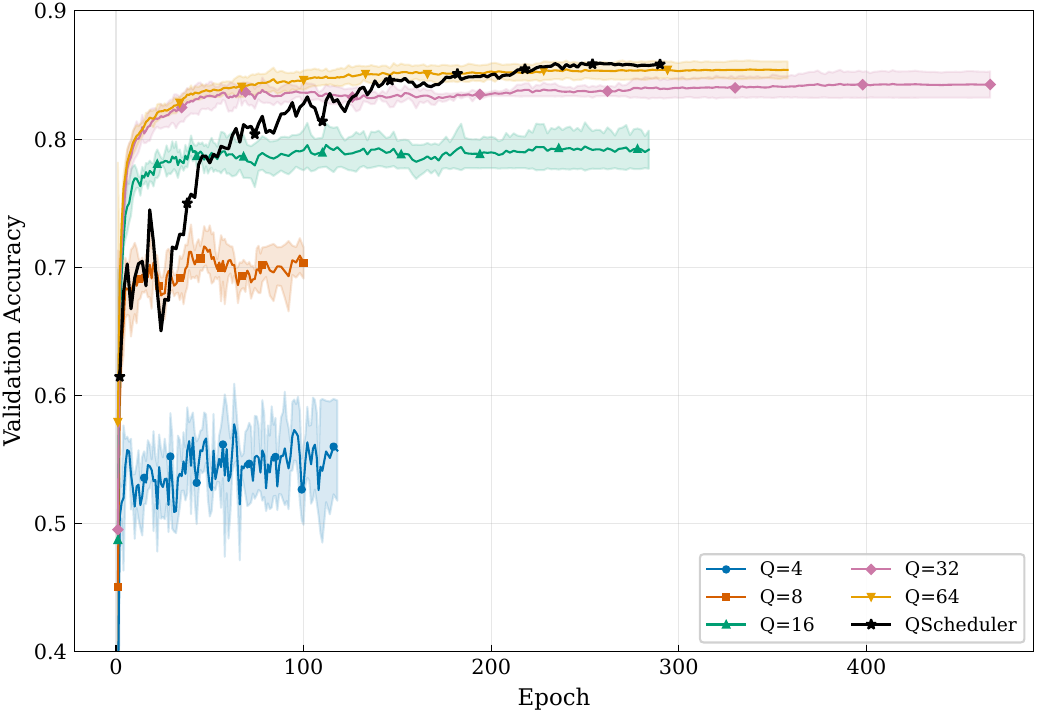}
\caption{Training curves for MobileNetV2 on EuroSAT: floating-point (top) and INT8 quantized (bottom). Lower $q$ values result in noisy gradients and early plateaus, with the effect more pronounced in INT8 due to the discrete weight space.}
\label{fig:mbnetv2_curves}
\end{figure}

Figure~\ref{fig:mbnetv2_curves} shows training curves for MobileNetV2 on EuroSAT in both floating-point and INT8 configurations. In floating-point (top), most configurations converge smoothly, with higher $q$ values achieving better final accuracy. QScheduler tracks the performance of mid-range fixed-$q$ values while adapting automatically. In INT8 (bottom), the discrete weight space introduces additional challenges: with too few samples ($q=4$), training is highly unstable and plateaus around 55\% accuracy. Increasing $q$ improves stability: $q=8$ reaches $\sim$70\%, $q=16$ reaches $\sim$80\%, and $q \geq 32$ converges to $\sim$85\%. QScheduler adapts to match the best fixed-$q$ performance without manual tuning.

\subsection{On-Device Experiment}

We validate our framework on the STM32N6 using a MobileNetV2 INT8 model fine-tuned on a subset of STL-10 (2000 images stored in Flash). The model is compiled via STEdgeAI and trained using our ZO implementation with gradient buffers in PSRAM.

\begin{figure*}[h]
\centering
\includegraphics[width=\textwidth]{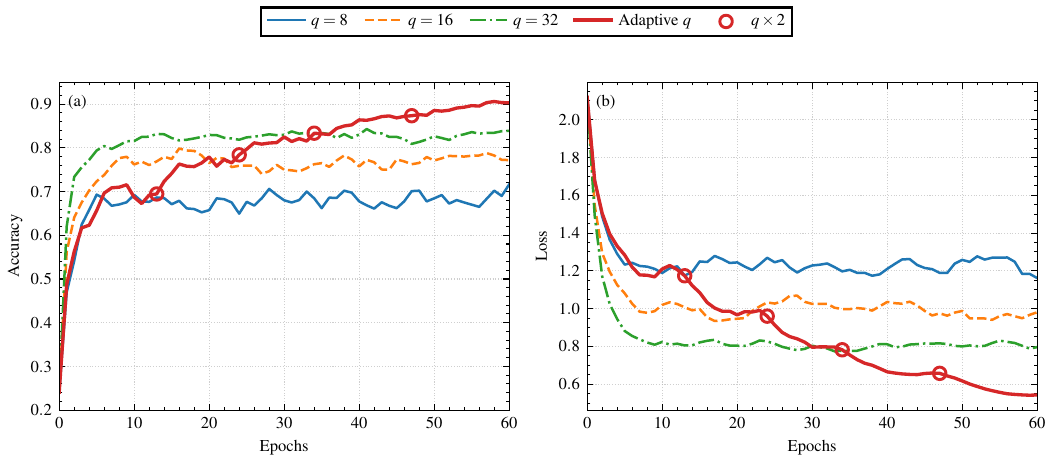}
\caption{QScheduler vs. fixed-$q$ baselines across training epochs on STM32N6. Dots indicate $q$ transitions.}
\label{fig:qscheduler_epochs}
\end{figure*}

Figure~\ref{fig:qscheduler_epochs} compares training curves for QScheduler against fixed-$q$ baselines. QScheduler begins with improvement at low $q$, then automatically increases $q$ when progress stalls (transitions marked by dots). By adapting gradient quality on demand, it matches the final accuracy of high-$q$ baselines.

\section{Discussion}
\label{sec:discussion}

The experiments reveal an important trade-off in ZO optimization: while higher $q$ values consistently improve accuracy, the marginal gains diminish beyond a certain point (e.g., $q=64$ to $q=1024$ yields only 1-2\% improvement). This suggests that beyond a certain point, the inherent approximation error of the ZO method (not the number of samples $q$) becomes the limiting factor. QScheduler naturally discovers this saturation point by monitoring progress, avoiding unnecessary computation once gradient quality is sufficient.

The accuracy gap between floating-point and INT8 quantized training (Tables~\ref{tab:accuracy_resnet18} and~\ref{tab:accuracy_mobilenetv2}) can be attributed to the compounding of two noise sources. In floating-point training, the ZO estimator introduces variance from finite sampling, but the perturbation magnitude $\mu$ can be chosen arbitrarily small. In INT8 training, each Rademacher perturbation $z \in \{-1, +1\}$ corresponds to a fixed step of magnitude $S_w$ (the quantization scale), imposing a lower bound on effective $\mu$. This discrete perturbation space amplifies gradient estimation noise, explaining why low-$q$ configurations degrade more severely in INT8. 

Additionally, the distributions of both weights and activations shift continuously throughout training. Despite the initial warmup phase providing a reasonable calibration baseline, the INT8 quantization scales remain fixed. As training progresses and these distributions drift, the fixed scales may no longer accurately represent the updated weight ranges, introducing additional quantization error that compounds with the gradient estimation noise and may further widen the accuracy gap with respect to floating-point training.

A practical advantage of QScheduler is its robustness when deploying to new scenarios. In real-world on-device learning, the optimal $q$ varies with the model architecture, dataset characteristics, and optimization space size. Determining this value through hyperparameter sweeps is often impractical on resource-constrained devices. QScheduler addresses this by automatically adapting gradient quality during training, requiring only conservative initial hyperparameters.

\section{Conclusion}

This work presented a framework for on-device learning on NPU-equipped microcontrollers using zeroth-order optimization. We introduced QScheduler, an adaptive algorithm that dynamically adjusts gradient estimation quality based on training progress, eliminating the need for manual hyperparameter tuning. Our experiments on EuroSAT demonstrated that QScheduler achieves accuracy comparable to well-tuned fixed-$q$ configurations across both floating-point and INT8 quantized training scenarios, without requiring exhaustive hyperparameter searches.

To the best of our knowledge, this work provides the first demonstration of on-device learning on the STM32N6 platform. Our results on the STM32N6 demonstrate that ZO optimization can enable training on NPUs designed for inference such as ST Neural-ART, extending on-device learning to a new class of embedded hardware.

\section{Future Work}

Several directions could improve the accuracy and efficiency of our framework. Dynamic scale recalibration during training could reduce the floating-point/INT8 accuracy gap, though at additional computational cost.

Regarding memory efficiency, the gradient accumulation buffer currently operates in FP16 format. Gradient quantization techniques (e.g., INT8 accumulation with periodic rescaling) could compress this buffer by $2\times$ or more, enabling training of larger models within the same memory constraints. Energy consumption and runtime profiling on the STM32N6 hardware are also planned as future work.

\section*{Acknowledgments}
This work was performed using computational resources from the “Mésocentre” computing center of Université Paris-Saclay, CentraleSupélec and École Normale Supérieure Paris-Saclay supported by CNRS and Région Île-de-France. The authors also acknowledge the use of Artificial Intelligence (AI), specifically Claude by Anthropic \cite{anthropic_claude}, for language refinement and assistance during the preparation of this manuscript.

\printbibliography

\end{document}